\documentclass[letterpaper]{article} 
\usepackage{aaai23}  
\usepackage{times}  
\usepackage{helvet}  
\usepackage{courier}  
\usepackage[hyphens]{url}  
\usepackage{graphicx} 
\usepackage{natbib}  
\usepackage{caption} 
\usepackage{algorithm}
\usepackage{algorithmic}

\usepackage{newfloat}
\usepackage{listings}
\DeclareCaptionStyle{ruled}{labelfont=normalfont,labelsep=colon,strut=off} 
\floatstyle{ruled}
\newfloat{listing}{tb}{lst}{}
\floatname{listing}{Listing}
\title{Mechanic Maker 2.0: Reinforcement Learning for Evaluating Generated Rules}
\author{
Johor Jara Gonzalez \textsuperscript{\rm 1},
Seth Cooper \textsuperscript{\rm 2},
Matthew Guzdial\textsuperscript{\rm 1}
}
\affiliations{
\textsuperscript{\rm 1} University of Alberta, Alberta Machine Intelligence Institute\\
\textsuperscript{\rm 2} Northeastern University\\
jaragonz@ualberta.ca, se.cooper@northeastern.edu, guzdial@ualberta.ca
}

\usepackage{bibentry}

\begin{document}

\maketitle

\begin{abstract}
Automated game design (AGD), the study of automatically generating game rules, has a long history in technical games research. 
AGD approaches generally rely on approximations of human play, either objective functions or AI agents. 
Despite this, the majority of these approximators are static, meaning they do not reflect human player's ability to learn and improve in a game. 
In this paper, we investigate the application of Reinforcement Learning (RL) as an approximator for human play for rule generation. 
We recreate the classic AGD environment Mechanic Maker in Unity as a new, open-source rule generation framework. 
Our results demonstrate that RL produces distinct sets of rules from an A* agent baseline, which may be more usable by humans.
\end{abstract}

\section{Introduction}

Procedural content generation (PCG) and procedural content generation via machine learning (PCGML) \cite{summerville2018procedural} are techniques used in video game research and development to automate the creation of game content. They provide developers with tools to create, test, and modify various game elements such as map generation, quest generation, and automated weapon design.

In video game development, game mechanics, also known as rules, are what determines how the game reacts to player input; they are defined as ``the rules, processes, and data at the heart of the game''~\cite{adams2012game} or ``methods invoked by agents designed for interaction with the game state''~\cite{sicart2008defining}. While designing balanced and engaging rules is a complex process that typically requires significant human expertise, there are some promising approaches to automated game design (AGD). In this approach, PCG approaches can be utilized to generate new rules, and deterministic autonomous agents can be employed to approximate human players, for example in Mechanic Miner~\cite{cook2013mechanic}. While most AGD approaches employ some form of human experience approximation to evaluate the generated rules, they mostly rely on static fitness functions or planners, which have certain drawbacks. Unlike these kinds of agents, humans do not just statically play games, particularly when confronted by new rules. Instead, they have to learn to play the games through trial and error.

We identify two main static analysis approaches used in generating and testing game mechanics in AGD. The first approach, taken by researchers such as ~\cite{togelius2008experiment}, and ~\cite{cook2012angelina}, generate rules based on heuristics for rule quality, without any approximation of human play. The second approach instead uses automated strategies to approximate human play and to test the various generated rules such as ~\cite{cook2013mechanic} and  ~\cite{isaksen2015exploring}. All of these approaches have employed search-based PCG, in which initial rules are generated then iteratively improved based on some fitness function. In the former case, this fitness function does not attempt to simulate gameplay, while in the latter cases, automated game playing techniques such as A* inform the fitness. To the best of our knowledge, Machine Learning (ML) approaches have not previously been applied to evaluating generated rules. 

We believe that A* agents may not be the best representation of human experience. One reason is that they can perform frame-sensitive movements that are beyond the capabilities of most humans. In addition, unlike humans, A* agents find an optimal route before moving, which means they do not react to game events in real-time like humans do. These agents are designed to make decisions based on a predefined set of rules, and their behavior is unchanging, unlike humans who can adapt to new situations and learn from experience. An alternative solution could be planners that included learning like the Learning Real-Time A* (LRTA*) algorithm \cite{korf1990real}. However, these approaches would still trend towards frame-sensitive, inhuman movements and, as we argue above, may not correspond well with human play. Further, as far as we know, no learning approaches have been applied to AGD. 

Reinforcement learning (RL) could provide a better approximation of human experience compared to A* and LRTA* because it requires an agent to learn how to interact with an environment.
This is more similar to the way a human might learn through trial and error.
However, unlike a human tester, an RL agent can be trained to evaluate a rule then reset (forgetting everything it has learned) to do the same for a variation on that rule.
RL agents have been used in the past to approximate human playtesting ~\cite{de2022automated}, but they have not yet been utilized for rule generation.

The purpose of this paper is to explore the potential of Reinforcement Learning (RL) for informing the fitness function in a search-based AGD approach. Specifically, we aim to generate game mechanics by employing a search-based PCG (SBPCG)~\cite{togelius2010search} approach with an RL-based fitness function. To accomplish this, we replicated the Mechanic Miner environment from ~\cite{cook2013mechanic}, a classic AGD environment. This study advances the field of game design by exploring new approaches to autonomous game design.

Our results demonstrate that a reinforcement learning (RL) agent-based fitness function led to more varied rules than an A* agent-based fitness function. This suggests machine learning algorithms have the potential to create unique game mechanics that may not be achievable through existing AGD methods.  Therefore, understanding the relative strengths and weaknesses of RL and A* algorithms is crucial for AGD researchers to continue to advance the field.

Our contributions for this paper include:
\begin{itemize}
    \item Applying RL as an approximator of human evaluation for rule generation via a search-based PCG fitness function.
    \item An open-source recreation of the Mechanic Miner environment in Unity where we generate rules and train our RL agent to use them.\footnote{\url{https://github.com/Harcurio/MechanicMiner}}
    \item A comparative analysis between RL and A* agents for rule generation in our environment.
\end{itemize}

\section{Related Work}

The history of AGD dates to roughly 1996~\cite{pell1996strategic}. Despite this almost thirty year history, AGD remains a relatively underexplored area of games research. 
AGD overwhelmingly relies on static analysis where the evaluator does not change or learn. Despite advances in machine learning, for the most part we haven't seen these approaches applied to AGD. 
The validation of game design, specifically the validation of generated game rules, has been approached in various ways. \cite{youtube} has proposed three different methods for evaluating the outputs of AGD. 

First, there is \emph{static analysis} which tries to evaluate things about the rules without playing the game directly, for example using heuristics of rule quality to inform a search-based generator. In this category, we can include works such as  ~\cite{pell1996strategic}, and  ~\cite{browne2010evolutionary}. In these approaches, unchanging heuristics were used to inform a search-based generator. While we also employ a heuristic, it is one informed by approximating human play via RL, as discussed further below.

Second, Cook identifies \emph{human evaluation} as a method to assess generated rules or systems to enhance their quality. \cite{guzdial2021conceptual} generated rules and evaluated them using human feedback, serving as a prominent example of this assessment approach. There are other examples of PCG research, such as  \cite{charity2020baba} and \cite{miller2022barriers}, that use human evaluation to assess content, but are not focused on rule generation. Although our work aims to evaluate generated rules in a more human-like manner, we do not rely on feedback from humans. The logistical challenges of collecting human feedback in a timely manner for rule generation would prove arduous and unproductive for both the participants and the research process.

Finally, Cook identifies \emph{agent-based playouts} as an approach to evaluate generated rules using AI agents. This category encompasses most of the AGD research to date because it allows a system to ensure new rules make a game playable and/or balanced. In these implementations, non-learning agents, like A* agents, planners, or random agents are used to inform a heuristic. Among the works that use an agent-based evaluation, we can include \cite{nelson2007towards}, \cite{treanor2012game}, \cite{kreminski2020germinate}, ~\cite{guzdial2018automated} and ~\cite{cook2016angelina,cook2022puck}. Perhaps most similar to our work ~\cite{sorochan2022generating}, employ Monte-Carlo Tree Search (MCTS), which can be classified as a model-based reinforcement learning (RL) approach, to evaluate generated game units with special abilities (rules). However, we make use of model-free RL, where the agent does not know the impact of a rule in order to better approximate how humans learn to use new rules to play games.

A distinct approach related to our work is PCG via Reinforcement Learning (PCGRL) as proposed by \cite{khalifa2020pcgrl}. In their work, an agent is used to transform random levels into final levels via reinforcement learning. To the best of our knowledge, PCGRL has not yet been applied to AGD. We do not categorize ur approach as PCGRL since we are not using RL as the generator. Instead, we employ RL as part of the fitness function to evaluate generated rules. We consider our work complimentary to PCGRL, as one could use PCGRL to generate rules instead of our SBPCG approach.

\section{System Overview}\label{sec:approach}

In this section, we describe our system in detail. Our objective was to evaluate the impact of an RL agent on evaluating the quality of generated rules. To generate rules we employed a Search-based Procedural Content Generation (SBPCG) approach \cite{togelius2010search} since it is the most common approach for AGD problems. To achieve this, we needed an environment to test the generated rules. We chose to reimplement Mechanic Miner \cite{cook2013mechanic} in Unity for this purpose. We made this choice as this would allow us to more easily compare to existing rule evaluation approaches for AGD. Due to the discontinuation of Adobe's support for Flash, we were unable to utilize the original Mechanic Miner environment implemented in Flixel. Instead, we leveraged the Unity ML-Agents framework to facilitate the integration of machine learning (ML) agents within Unity. By employing a deep reinforcement learning approach, we successfully developed an agent capable of learning to navigate our recreated Mechanic Miner environment. This was achieved by utilizing generated rules obtained through our SBPCG approach. This RL agent then gives feedback to our SBPCG approach via a fitness function to approximate rule quality.

\subsection{Mechanic Maker 2.0 Environment}
\begin{figure}
    \centering
    \includegraphics[width=8cm]{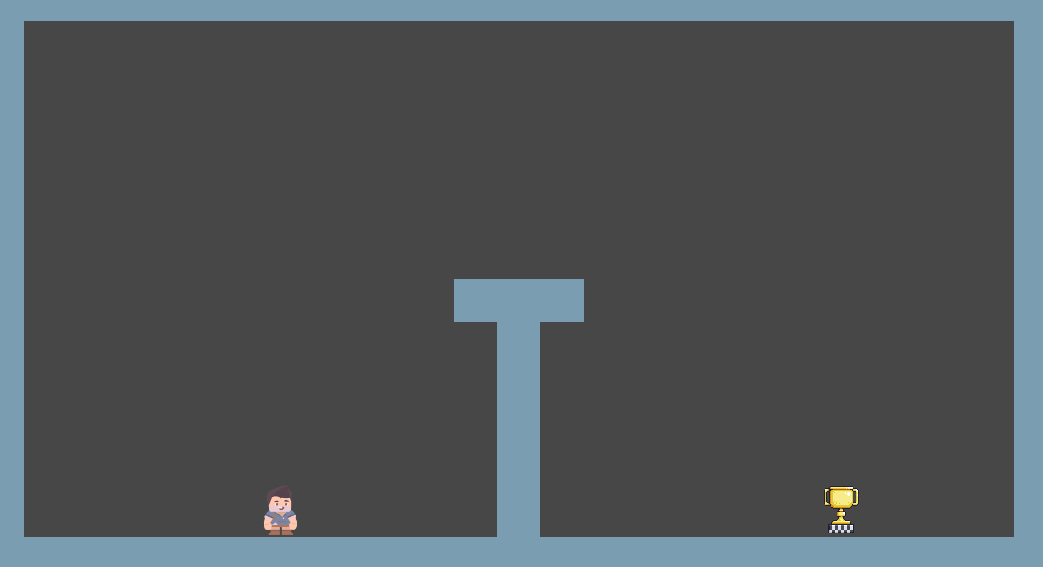}
    \caption{Our recreation of Mechanic Miner in Unity}
    \label{fig:env}
\end{figure}

Our 2D platformer Unity project employs a grid-based setup and reflects the original Mechanic Miner environment. Mechanic Miner utilized the Flixel library for ActionScript 3, which is no longer supported. We replicated the Mechanic Miner evaluation environment design in Unity as seen in Figure~\ref{fig:env}. This is a simple environment, with a player that always starts on the left, a goal (a trophy in this case) on the right, and a `T' shaped obstacle in between them.

One of the defining features of the original Mechanic Miner was the use of code reflection, which gave the search-based rule generator access to all public variables. This was thanks to the Flixel game engine, and is not directly supported in the Unity game engine. Because of this, in our approach, we replicate this feature by making all the variables related to movement public, allowing our generator to output rules that refer to and modify these variables. 

As with the original Mechanic Miner, there are a set of initial player movement rules present in the environment prior to any rule generation. The initial movement rules in this environment are moving to the left, moving to the right, and jumping. We cover the associated variables below. Notably, as with the original Mechanic Miner, with only these initial rules, it is impossible for the player to reach the goal in the environment. Specifically, the jump height is insufficient to reach the top of the `T' shape obstacle, and it's not possible to otherwise move past it due to collisions. Therefore, agents have to rely on generated rules to reach the goal.

We will now provide a more detailed explanation of the public variables associated with movement in our Mechanic Maker 2.0 environment, which were available for our generator to draw on to construct the generated rules:
\begin{itemize}
    \item  \textbf{JumpForce}, which determines the force applied to the jump and affects the player's jump height. This is carefully tuned such that it is only possible to jump to the midpoint of the `T' shaped obstacle with the initial jump movement rule. 
    \item  \textbf{Speed}, which is the maximum movement speed of the player on the ground. This also determines the horizontal movement speed while jumping, multiplied by $0.65$ to represent air friction.
    \item  \textbf{Position.X}, which refers to the player's horizontal position in the environment.
    \item  \textbf{Position.Y} which refers to the player's vertical position in the environment.
\end{itemize}

We could have drawn on additional variables, such as the player's scale, collision box, or gravity, or included a separate air friction variable. However, our purpose for this initial study was to compare RL and an existing generated rule evaluation approach (A*). As such, additional variables would have complicated our results. We leave investigating the impact of additional public variables for future work. 

\begin{table}[t]
\centering
\resizebox{\columnwidth}{!}{%
\begin{tabular}{|c|c|c|c|c|}
\hline
Variables & Comparators & Comparison & Effect & Effect \\
 &  & Values &  & Values \\ \hline
speed & \textgreater{} &  & add &  \\
jumpForce & \textless{} & [1,20] & substract & [1,10] \\
position.X & == &  & multiply &  \\
position.Y &  &  & divide &  \\
\multicolumn{1}{|l|}{} & \multicolumn{1}{l|}{} & \multicolumn{1}{l|}{} & residue & \multicolumn{1}{l|}{} \\ \hline
\end{tabular}%
}
\caption{The possible variables, comparisons, effects, and values that could be drawn on by our search-based PCG (SBPCG) rule generation system. These possible values and ranges define our search space.}
\label{tab:variables}
\end{table}

Our rule generation system's objective is to generate a new rule that will facilitate the agent's ability to reach the goal located on the other side of the T-shaped obstacle. To achieve this, we ran a total of five agents at the same time in parallel, all agents were synchronized to use the same rule through a file sharing system. 

We employed a fixed number of 2,000 training episodes per agent, thus it is likely most agents will reach the goal multiple times while training. Each time the agents reached the goal, we saved information on the specific training session. To allow for later analysis of the agent's training process we saved the following information:
\begin{itemize}
    \item \textbf{Generated rule}, the rule that has been tested during this training process. 
    \item \textbf{Time to train}, refers to the total time it took the whole system to train.
    \item \textbf{Time to goal}, refers to an array that updates each time an agent reaches the goal.
    \item \textbf{Goal completion}, a count of the number of times agents reached the goal and updates every time an agent reaches the goal during training.
    \item \textbf{Deaths}, there are no harmful elements in the environment, but we assume the agent instantly dies if it leaves the environment's boundaries. Thus, this tracks how many times agents moved outside the boundaries of the environment during training.
    \item \textbf{Actions taken}, records the number of actions taken by the agent each time it reaches the goal during a training cycle, along with the specific sequence of actions the agent took.

\end{itemize}

\subsection{RL Agent}

In the original Mechanic Miner, a breadth first search (BFS) agent evaluated whether newly generated rules would allow an agent to reach the goal. In our approach, we employed an RL agent in the same role instead. Despite using a learning approach there are some commonalities that support this decision. One aspect is the ability to achieve deterministic behavior.\footnote{by setting the random seed on the RL agents.}
Our RL Agent trains based on the following Markov decision process (MDP) :
\begin{enumerate}
    \item States: We utilized two 2D raycasts surrounding the agent as a frontier to limit the agent's perception of the environment. While one raycast only detects platforms below the player, the other raycast can sense objects such as the goal and walls, and their x,y positions. 
    The returned values of these two raycasts define the state.
    Our initial state is an agent in the air slightly above the ground as seen in Figure \ref{fig:env}.
    \item Actions: We have a total of five possible actions the agent can take in each state, which we represent in a one-hot encoding: \[Move left, Move right, Jump, New rule, Nothing\]
     
    \item Transition function: We assume deterministic actions, with the agent having perfect control of its state transitions outside of gravity and collisions in the environment. 
    \item Rewards: The reward of the agent is given by the following equation:
    \begin{equation}
        \sum  (Move_L + Goal_R - \frac{1}{Max_{ST}})
    \label{equ:Reward}
    \end{equation}
    \begin{itemize}
        \item $Move_L$ is a positive reward given for each movement made during the episode.
        \item $Goal_R$ is an arbitrarily large reward for reaching the goal.
        \item $Max_{ST}$ is a penalty incurred for every frame of the training episode.
    \end{itemize}
\end{enumerate}

Since the agent had no prior knowledge of the environment, we trained it from scratch, meaning that it had to learn how to use the default rules of the environment and any new rule every time.

We used Proximal Policy Optimization (PPO) \cite{schulman2017proximal} as the reinforcement learning algorithm for our agent, given that it is ideal for training agents that do not have optimal actions to reach a goal. We drew on the default PPO implementation in Unity ML Agents \cite{juliani2018unity}. 
To expedite the learning process, we trained five agents in parallel, each with the same configuration. Using this method allowed us to use the experience of each agent to train the artificial neural network in a more efficient manner. We also made two changes to the configuration parameters of the RL agent:
\begin{itemize}
    \item Modifying the time horizon from the default value of 32 to 64, which corresponds to the amount of experience collected per agent.
    \item Adjusting the beta value from the default of 0 to 1e-3 to encourage agent exploration.
\end{itemize}

\subsection{Search-based Generation of Rules}

We employed a search-based PCG (SBPCG) process, given its popularity in AGD research \cite{togelius2010search}.
In our search process the first step was to randomly generate four different rules using the properties listed in Table \ref{tab:variables}. We selected random values for each available variable. Once we generated the initial population of four rules, we trained the RL agent for each rule and saved the results. These results informed our fitness function. Though we refer to our evaluation function as fitness function, a lower value is preferred. In the case that the Rl agent was unable to learn to reach the goal with any of the generated rules, we repeated the first step: generating four new rules. Once we found a set rules where at least one could reach the goal, we selected the best rule based on our fitness function (below). This became our first current rule, and from here we began a stochastic greedy search to further optimize this current rule. Until we converged, we iterated through four random neighbors of our current rule. To generate neighbor rules, we randomly modified a variable, other than the name variable in Table~\ref{tab:variables}. A value of one was added or subtracted at random to the value of the effects or the value of the comparator. We evaluated all the neighbors based on the fitness function below and selected the best among them and the current rule. This process repeated until we converged. Due to the stochastic nature, we consider convergence to require $3$ repetitions of neighbours that do not outperform the current rule.

In comparison to our approach, the original Mechanic Miner employed a genetic algorithm that generated large populations of rules. We chose to employ a much simpler search-based approach in the form of a stochastic greedy search for two main reasons. First, RL agents take much longer to train than planners do to plan, meaning that each evaluation by our fitness function is much more expensive. Second, we wanted to determine the difference in generated rules when using an RL agent and a more traditional planning approach for evaluation. Thus, we felt a simpler search algorithm would allow us to quickly identify and compare the relative local optima. 

\begin{equation}
\resizebox{.8\hsize}{!}{$T_{G} + (1- \frac{|T_{M} - N_R|}{Max(T_M, N_R)} )* S\ - O_Z$}
\label{form:2}
\end{equation}

In equation \ref{form:2}, we present our fitness function for the rules evaluated by the RL agent. In this function, we use several variables to evaluate the agent's performance. Below we define each variable:

\begin{itemize}
    \item $T_{G}$: Represents the number of steps taken to reach the goal. If the goal was not reached, its value is 0.
    \item  $T_{M}$: Signifies the total number of times the agent used the base rules during training.
    \item $N_R$: Indicates the total number of times the new rule was utilized during training.
    \item $S$: An integer used to assign weight to the usage of rules.
    \item $O_Z$: Represents the total number of times the agent went out of bounds.
\end{itemize}

The function aims to determine the agent's fitness by assigning a score. It returns the maximum value when the agent successfully reaches the goal and uses the new rule approximately the same number of times as the other rules. A lower fitness function is preferred since we only analyzed rules that were able to reach the goal. We have designed it this way to incentivize the usage of the new rule and the basic ones at a roughly equal rate. This encourages the integration of the new rule with the existing ones, and attempts to avoid cases where an agent only depends on a new rule.


\section{Evaluation}  

Our study aims to investigate potential differences between using a dynamic RL agent and the more traditional AGD approach of static planners, such as A* agents, for evaluating generated rules. The goal is to identify any distinctions between these methods in the context of rule evaluation. It is important to note that our primary focus is not on optimizing the agents' processing time, but rather on testing the impact of any differences in their evaluation of the newly generated rules. Given that we focused on testing this new approach (RL applied to AGD evaluation), we decided not to incorporate a human subject study in this initial stage.

To establish a baseline, we implement an A* agent and compare its performance to our RL agent when it instead informs the fitness of our search-based PCG rule generator. We generate rules and compare them qualitatively in terms of how they change a player's movement, as well as quantitatively by measuring the differences between them.
Our primary objective is to identify and characterize any differences between the rules generated by these approaches, thus this evaluation method is a natural choice to achieve this goal.

\subsection{A* Agent}

To compare the performance of an A* agent and an RL agent, we developed an A* agent to simulate the Mechanic Miner approach. We employed an A* agent even though the original Mechanic Miner paper used a breadth first search (BFS) agent because their BFS agent made use of a closed set and prioritized shorter paths, making it equivalent to an A* agent \cite{cook2013mechanic}. In addition, A* agents have been employed in other instances of AGD research \cite{guzdial2018automated}. To ensure a fair comparison, we employed a set of actions that closely resembled those used by the RL agent. Our heuristic is both admissible and consistent, ensuring optimal A* paths. The A* agent endeavors to reach the goal by simulating one action at a time, meticulously searching for the shortest path between the start and the goal using the generated rules, if such a path exists. Similar to the RL agent's reward function, the heuristic of the A* agent is based on the distance to the goal.

\begin{equation}
\resizebox{.9\hsize}{!}{$
    T_G + (1- \frac{|T_M - N_R|}{Max(T_M, N_R)} )*  S + (-1 *D_G)
$}
\label{form:3}
\end{equation}

In equation \ref{form:3}, we present our A* fitness function, which is designed to closely approximate the RL fitness function. Outside the fitness function, the two AGD SBPCG approaches are identical. Below we describe the variables of the A* fitness function in detail:

\begin{figure}[tbh]
    \centering
    \includegraphics[width=8cm]{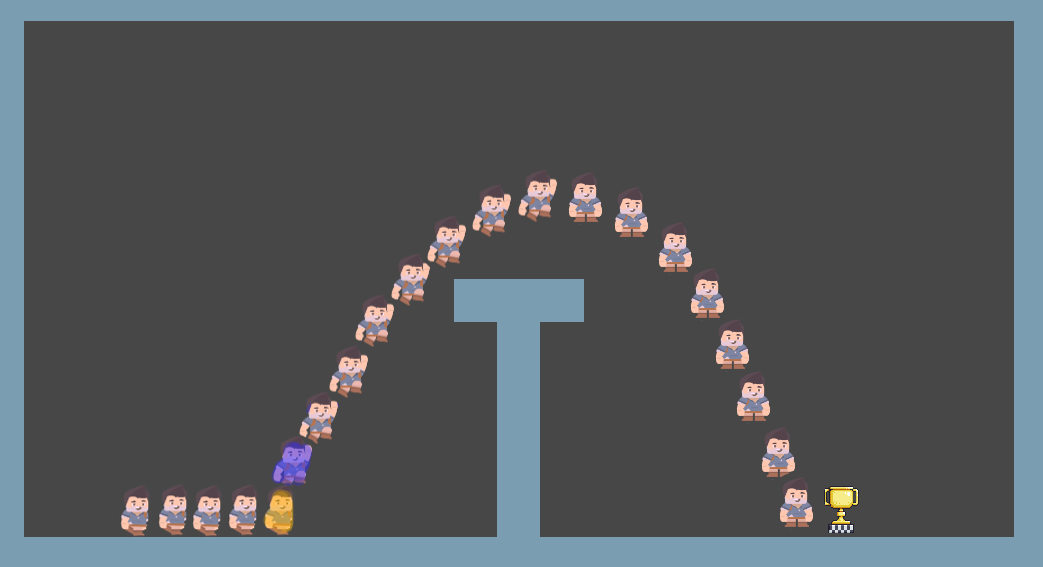}    
    \caption{This image represents the rule where the jumpForce value is modified, RL3. The yellow avatar indicates when the rule was triggered and the purple avatar is roughly equivalent to the next frame.}
    \label{fig:jumpFrames}
\end{figure}

\begin{itemize}
    \item $T_G$: Represents the number of time steps taken to reach the goal. If the goal was not reached, its value is 0.
    \item $T_M$: Signifies the total number of movements explored during planning with the A* agent.
    \item $N_R$: Indicates the total number of times the new rule was used during planning.
    \item $S$: Represents a weight assigned to the number of used rules, identical to the variable in the RL fitness function.
    \item $D_G$: Denotes the distance of the agent from the goal at the end of the path.
\end{itemize}

The underlying implication of this fitness function is to prioritize lower fitness values. This preference arises from our analysis, which focused only on rules capable of reaching the goal. We noticed that shorter or simpler paths may lead to worse rule evaluation compared to more complex rules. However, the fitness function still maintains a balance by ensuring that the new rule is employed roughly as frequently as the other rules. The primary goal of both fitness functions remains to favor agents that successfully reach the goal while considering the impact of different rule usage. 

As part of our effort to further ensure a fair comparison between our approach and the AI agent baseline, we implemented a variation that employed random action selection when confronted with the predicament of two equally valuable paths. Although we introduced this modification, we observed that the output rules remained unaltered. We suspect this is due to the fact that our A* agent was still finding optimal paths, so the evaluation of the generated rules remained unchanged. In another attempted variation, we developed a greedy agent but discovered that it output only a subset of the same rules identified by the A* agent. Therefore, we only include results from the A* agent approach.


\begin{figure}[tbh]
    \centering
    \includegraphics[width=8cm]{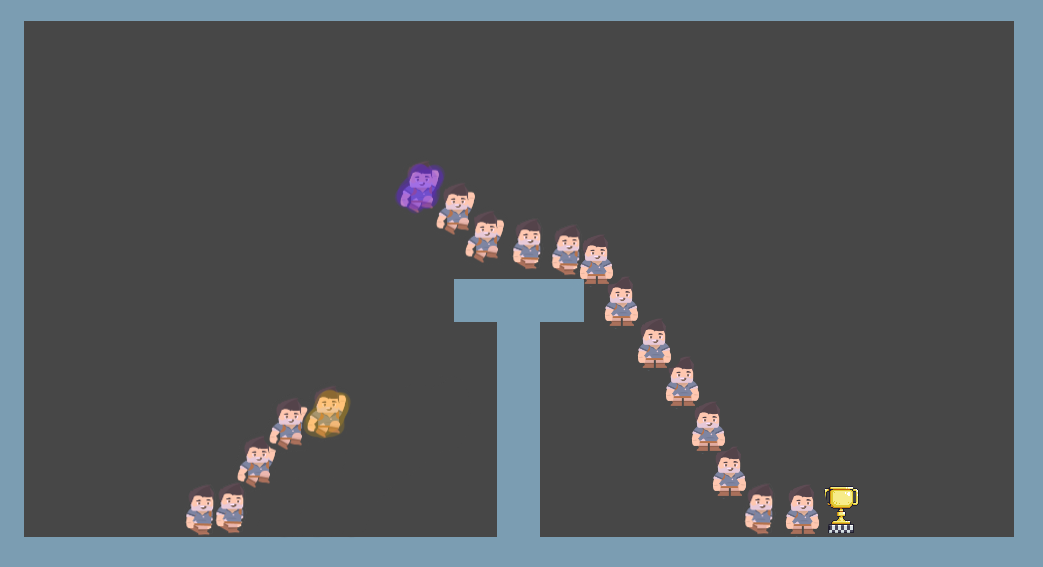}    
    \caption{This image depicts one of the different ways that the RL agent utilized one of the generated rules RL2. In this case, the agent needed to perform a jump action before triggering the rule, indicating that the rule was dependent on the agent's previous action. The yellow avatar indicates when the rule was triggered and the purple avatar is the next frame}
    \label{fig:yframes}
\end{figure}

\begin{figure}[tbh]
    \centering
    \includegraphics[width=8cm]{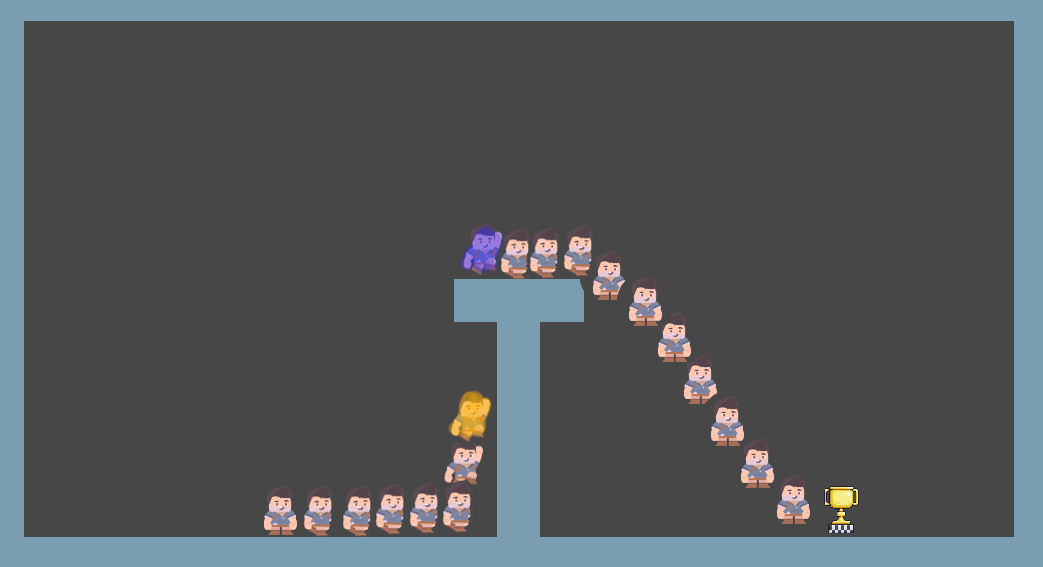}    
    \caption{This image represents a different way that the RL agent used the same rule shown in Figure 3, RL2. We found that similar rules found by the A* agent exhibited the behavior depicted in this Figure but not the one from Figure 3. The yellow avatar indicates when the rule was triggered and the purple avatar is the next frame}
    \label{fig:yframesA}
\end{figure}

\begin{figure}[t]
   \centering
    \includegraphics[width=8cm]{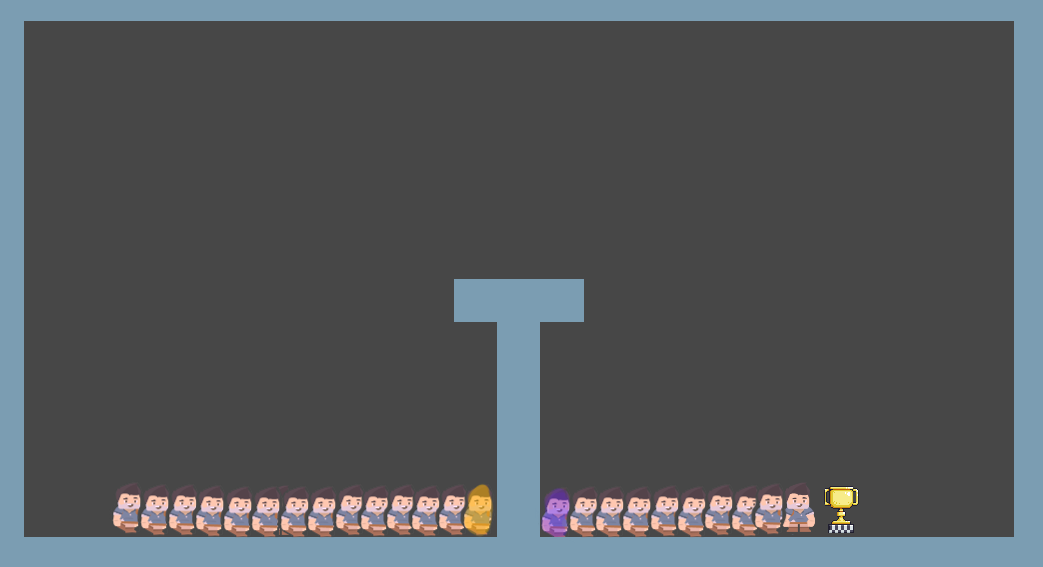}    
    \caption{This image represents rule RL4 evaluated by the RL agent. In this rule, the agent used a speed increase to skip the obstacle. The yellow avatar indicates when the rule was triggered and the purple avatar is the next frame}
    \label{fig:Run}
\end{figure}

\section{Rule Description}

In this section, we describe the rules generated using both approaches. Each time we ran the environment, we obtained a rule from each approach. With the A* agent-based generator, we generated a total of 12 rules before reaching a point where the generator stopped outputting unique rules. On the other hand, the RL agent-based generator produced 20 rules before it also ceased to generate unique rules.

Given the different number of unique outputs, we selected a subset of five from each agent to discuss further. In terms of the selection, we select from the A* agent baseline's rules at random due to their relative uniformity. 
Due to the variety of RL agent rules, we instead selected one of the rules for each of the variable effect types at random. 
We include these rules obtained by our search-base PCG approach evaluated by the RL agent in Table \ref{tab:rlagent}, and those evaluated by the A* agent in Table \ref{tab:astart}. 
Despite the different selection criteria here, all generated rules were used for quantitative evaluations below.

\begin{table}[t]
\centering
\resizebox{\columnwidth}{!}{%
\begin{tabular}{|c|c|c|c|c|c|}
\hline
RL Agent & Name & Comparator & Value & Effect & Value \\
Rule & Variable &  & Comparator &  & Effect \\ \hline
RL1 & speed & == & 20 & multiply & 4 \\ \hline
RL2 & position.y & \textgreater{} & 12 & add & 7 \\ \hline
RL3 & jumpForce & \textgreater{} & 3 & add & 10 \\ \hline
RL4 & speed & \textless{} & 10 & add & 8 \\ \hline
RL5 & position.x & \textless{} & 10 & multiply & 3 \\ \hline
\end{tabular}%
}
\caption{A selection of rules obtained by the RL agent chosen semi-randomly from the outputs generated during the experiments. This selection aims to showcase the different types of rules that can be generated, providing insights into the variety of solutions the RL agent can produce.}
\label{tab:rlagent}
\end{table}

\begin{table}[t]

\centering
\resizebox{\columnwidth}{!}{%
\vspace{-15pt}
\begin{tabular}{|c|c|c|c|c|c|}
\hline
A* Agent & Name & Comparator & Value & Effect & Value \\
Rule & Variable &  & Comparator &  & Effect \\ \hline
A*1 & jumpForce & \textless{} & 11 & add & 10 \\ \hline
A*2 & jumpForce & \textgreater{} & 3 & multiply & 10 \\ \hline
A*3 & jumpForce & == & 4 & add & 9 \\ \hline
A*4 & position.y & \textless{} & 8 & add & 8 \\ \hline
A*5 & position.y & \textgreater{} & 3 & multiply & 2 \\ \hline
\end{tabular}%
}
\caption{A selection of rules acquired by the A* agent, chosen at random from the outputs generated during the experiments, offering insights into the types of solutions that can be generated by the A* agent.}
\label{tab:astart}

\end{table}

To better understand how the rules work, we walk through an example. ``$PosY<8+9$'' means that if the player's position is less than 8 on the Y axis, 9 will be added to the `Position.Y' variable and the player will be moved to that position if the rule is triggered. Note that both agents had to choose to use a generated rule, we did not generative any passive or automatic rules.

We find intriguing differences when comparing the performance of the RL and A* agents within our environment. Notably, we found that the A* agent demonstrated faster goal-reaching capabilities during training sessions, while the RL agent required more time to train, typically ranging from 5 to 15 minutes per session. This is not surprising given the differences between training and planning. It is worth reiterating that our primary focus in this experiment was not achieving the fastest solution to the problem at hand.
In addition to the contrasting training times, we discovered that the evaluation by the RL agent led to the emergence of more varied rules, utilized in a wider variety of ways. We delve into some of these findings below. Furthermore, we observed a similarity in the rate of use of new rules compared to the old rules between the RL agent and the A* agent.

Only one of the rules (rule RL4 in Table \ref{tab:rlagent}) evaluated by the RL agent allowed the agent to pass through the wall by increasing its speed. We visualize this in Figure~\ref{fig:Run}. This was an unusual behavior that the A* agent-based approach did not replicate. Conversely, the A* agent was more consistent in finding solutions that were less surprising to us and did not evaluate rules with similar effects like passing through walls. We anticipate this was due to the greater exploration associated with learning.

Another interesting observation was the use of rules where the position.y variable is changed, like rule RL2 of the RL agent-based approach. This observation suggests that the RL agent was able to explore and exploit the environment in ways that the A* agent could not. The A* agent-based approach led to rules that could help the A* agent achieve the shortest path to the goal as seen in Figure \ref{fig:yframesA}. whereas the RL agent was able to lead the generator to discover different ways of reaching the goal  Figure \ref{fig:yframes} with similar rules.

While the A* agent-based approach was able to consistently lead to rules with jumpForce as the main variable, such as rules A*1-A*3 in Table~\ref{tab:astart}, we observed that the RL agent-based approach had difficulty with this type of rule. By difficulty, we mean this type of rule was much less commonly output, but that it was not impossible as in the case of RL3. Our hypothesis is that since the agent must trigger this kind of rule at a specific position, it is comparatively easier for the A* agent to accomplish this task given the lack of stochastic actions.
During the training of the RL agent, we discovered that it was able to circumvent our fitness function, which was designed to ensure that the number of movements between different actions was balanced. Our intention was to prevent the agent from using any rules that would simply teleport it directly to the goal. However, we observed that the RL agent was able to exploit this fitness by first moving away from the goal, via random movements, and then using a rule that enabled it to reach the goal as in the cases of RL4 and RL5.
This observation was intriguing because it revealed the agent's ability to adapt and find `creative' solutions to achieve its objective. However, it also highlighted the limitations of our fitness function and the need for more sophisticated metrics to evaluate the generated rules accurately. These findings demonstrate the complexity of training an RL agent for rule evaluation and the importance of improvement and iteration in this process.

\begin{table}[t]
\centering
\begin{tabular}{|c|c|c|}
\hline
 Rule Number & RL Agent  \\ \hline
RL1 & 42\%  \\ \hline
RL2 & 9.62\%  \\ \hline
RL3 & 36\%  \\ \hline
RL4 & 78.2\%  \\ \hline
RL5 & 11.58\% \\ \hline
Mean $\pm~\sigma$ & $35.48 \pm 27.9$ \\ \hline
\end{tabular}%
\vspace{1pt}
\caption{ Percentage of new rules used when training the RL agent}
\label{tab:goalRL}
\end{table}

\begin{table}[t]
\centering
\begin{tabular}{|c|c|c|}
\hline

 Rule Number & A* Agent \\ \hline
A*1 & 51.70\% \\ \hline
A*2 & 57.9\% \\ \hline
A*3 & 54.01\% \\ \hline
A*4 & 32.48\% \\ \hline
A*5 & 21.3\% \\ \hline
Mean $\pm~\sigma$ & $43.4 \pm 15.8$ \\ \hline
\end{tabular}%

\caption{Percentage of the new rules used when planning with the A* agent}
\label{tab:goalA}
\end{table}

\begin{table}[t]
\centering
\vspace{-10pt}
\begin{tabular}{|c|c|c|}
\hline
RL vs RL & A* vs A* & RL vs A* \\ \hline
206.20 & 261.0 & 208 \\ \hline
\end{tabular}%

\caption{This table shows the similarity between the generated rules, where the max value is 396 which means all the rules are identical, and lower values would mean less similarity between them. This table was made by selecting 12 random rules from each approach's output.}
\label{tab:sim}
\end{table}

\begin{figure}[t]
    \centering
    \includegraphics[width=8cm]{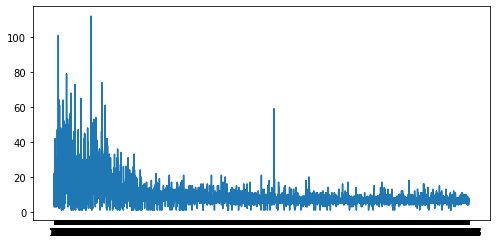}    
    \caption{This image plot represents the use of RL4 by the RL agent during training, we can see that the rule is more often used at the beginning of the training process. Where the Y-axis represents times the rule was used per cycle of training, and X-axis represents the training cycles.}
    \label{fig:training}
\end{figure}

\subsection{RL vs A*}

To further compare the output rules between our two evaluators, we calculated the percentage of times each rule was used during RL agent training and during A* agent planning. The results are presented in Table \ref{tab:goalRL} and Table \ref{tab:goalA}, respectively. It is worth noting when attempting to understand these results that both fitness functions preferred new rules that were used in a similar percentage to the other possible actions. Since we have 5 actions, the best result should be 20\% of the total usage. The results show that some of the rules selected by the RL agent tend to be used the same amount as the other actions, while the rules obtained by the A* agent tend to be used more than the other actions.
We anticipate this is due to the A* agent relying on the ``powerful'' rules more in the optimal paths it finds.
Notably for the RL agent we also have to consider how the rule usages changed over time.
Since the RL agent needs to be trained to use the rules correctly, it uses the rules randomly more often at the beginning of the training process. However, towards the end of training, it uses the rules more consistently. For example, we can see this trend in Figure \ref{fig:training} for RL4.

In addition to assessing the frequency of rule usage, we also examined the diversity of rules generated by each approach. To accomplish this, we categorized the rule variables into symbolic and numeric categories. The symbolic variables were the name variable, comparator, and effect, while the numeric variables were the value comparator and value effect (see Table \ref{tab:variables}). Subsequently, we devised a similarity function that aggregates similar values and normalizes the numeric categories. Specifically, for each symbolic variable, a value of 1 is assigned when two rules share the same symbolic variables, and the same principle applies to the numeric variables. In cases where the numeric values do not match exactly, a normalized difference of 1 is subtracted.


With our similarity function, we can compare the output rules in a summative manner.
In Table \ref{tab:sim} we use this similarity function to approximate ``similarity'' values across sets of rules. This basically acts like an inverse distortion measure for clustering. We measured the similarity in a pool of rules to all other rules in another pool. 
Due to the differences in the number of output rules, we randomly selected 12 output rules from each agent. 
RL vs RL thus gives us the similarity across the RL rules, A* vs A* gives us the similarity across the A* rules, and RL vs A* gives us the similarity between the two pools.  
In Table \ref{tab:pers}, we summarize these output rules in terms of which variable the output rules altered. As is obvious from these results, the A* agent-based approach had far less variety in its generated rules compared to the RL agent-based approach.
In addition, we found that the RL agent's rules were about as dissimilar to each other as they were to the A* rules. Based on the results, it appears that the RL agent-based approach is better at discovering rules that are more distinct from each other compared to the A* agent-based approach, which follows from the results shown in Table \ref{tab:pers}. 
\begin{table}[t]
\vspace{2pt}
\centering
\resizebox{\columnwidth}{!}{%
\begin{tabular}{|c|c|c|c|c|}
\hline
 & jumpForce & speed & position.X & position.Y \\ \hline
A* & 26\% & -- & -- & 74\% \\ \hline
RL & 15\% & 45\% & 5\% & 35\% \\ \hline
\end{tabular}%
}
\caption{The percentages of the different variables used as part of the effect for each generator. The first row gives the percentages for the A*
agent-based SBPCG rule generator, which only ever output rules changing the agent’s jumpForce and position.Y variables. The second row gives the percentages for the RL agent-based SBPCG generator, which had a wider spread of rules.}
\label{tab:pers}
\end{table}

\section{Limitations and Future Work} \label{sec:future}
Our observations highlight the strengths and weaknesses of each agent for rule evaluation, and the potential implications of these findings for AGD approaches and PCG system evaluations. While the RL agent has the potential to help find unconventional solutions to problems, it is slower and may find ways to ``hack'' a fitness function and the encoded rule quality concepts. In contrast, the A* agent is faster and more consistent in finding standard solutions but may struggle to find more innovative ones, without being forced to explore via a more stochastic or diversity-driven approach.

We believe that our system could be improved to find more interesting rules. One potential change is to add more variables or modify the ranges of allowable changes, which could increase the variety of generated rules. This would require careful consideration of the range of values that the variables could take, and their relationship to the other variables in the system. We could also consider adding new variables that are not part of the player but impact it, such as gravity, friction, and so on, which could further increase the diversity of generated rules.

We could further increase rule diversity by allowing rules to be activated by the value of another variable, creating more complex and nuanced rules. This would require developing a new algorithm for rule generation that could handle this additional complexity, but could lead to new types of rules. We could also consider conditional rules, which would trigger only if certain conditions were met, such as the presence of a specific value in a given variable.

Further exploration on the generation side could also improve the output. For example, we could explore a quality-diversity approach instead of a greedy search. This approach may lead to more diverse and creative solutions and could allow an A* agent-based approach to output more innovative rules. However, we felt a greedy approach was most effective for comparing the local maxima identified by two different fitness functions.

Currently, we use an RL agent as part of our rule evaluation process. However, we could investigate using Reinforcement Learning solely for rule generation or for simultaneous rule generation and evaluation. This would require retraining either a single RL agent on a larger and more diverse environment or two different RL agents, which could be time-consuming and computationally intensive. Although training times may increase, this would allow for an end-to-end system that could configure the rules during training to better meet an RL agent's needs. To maximize the benefits of using the RL agent, we could also look into ways to improve its training time, such as using transfer learning or other optimization techniques.

Finally, we could consider revising our fitness function, taking into account more parameters to try to avoid the ability for the RL agent to `hack' it. By doing so, we could more accurately evaluate the quality and utility of the generated rules. However, this would require significant experimentation and testing to determine the optimal set of parameters to use, and how they should be weighted to balance accuracy and computational efficiency.

In the future, we hope to help support game designers and developers in terms of not just generating new rules, but rules that specifically fit into an existing set of mechanics. This was part of the motivation for the design of our fitness function. We imagine a system that could suggest new, complimentary mechanics to game designers and developers during production, similar to the work of Machado et al., but without requiring an existing dataset of rules \cite{machado2019evaluation}.\\

\section{Conclusions}

In this paper, we investigated the application of RL to the evaluation of generated rules as part of a search-based PCG AGD system. We recreated the classic Mechanic Miner environment in Unity, setup to work with the Unity ML agents library. We ran a comparison between our RL agent-based approach and a more traditional A* agent-based approach. We found that our RL-agent based approach was able to output a wider variety of rules, including cheating its way past our fitness function. We hope that this work, through our open source recreation of Mechanic Miner and our novel results, invigorates the AGD research field, inspiring new approaches for generating and evaluating rules.

\section*{Acknowledgments}

We would like to acknowledge Pixel Frog and Kenney, who created the trophy sprite and all other assets in the paper, respectively. We would also like to thank Dr. Michael Cook for allowing us to use the ``Mechanic Maker'' name. This work was funded by the Canada CIFAR AI Chairs Program, Alberta Machine Intelligence Institute, and the Natural Sciences and Engineering Research Council of Canada (NSERC).

\bibliography{aaai23}

\end{document}